\def\paperlanguage{}
\pdfoutput=1 

\documentclass[letterpaper, 10 pt, conference]{ieeeconf}  

\usepackage{bm}
\usepackage{cite}
\usepackage{flushend}
\include{preamble}

\IEEEoverridecommandlockouts                              

\title{\LARGE \textbf
  {
    \switchlanguage%
    {%
      Hardware Design and Learning-Based Software Architecture of Musculoskeletal Wheeled Robot Musashi-W\\for Real-World Applications
    }%
    {%
      実世界応用に向けた筋骨格車輪型ロボットMusashi-Wの\\身体設計と学習型ソフトウェアアーキテクチャ
    }%
  }
}

\author{Kento Kawaharazuka$^{1}$, Akihiro Miki$^{1}$, Masahiro Bando$^{1}$, Temma Suzuki$^{1}$, Yoshimoto Ribayashi$^{1}$,\\Yasunori Toshimitsu$^{1}$, Yuya Nagamatsu$^{1}$, Kei Okada$^{1}$, and Masayuki Inaba$^{1}$
  \thanks{$^{1}$ The authors are with the Department of Mechano-Informatics, Graduate School of Information Science and Technology, The University of Tokyo, 7-3-1 Hongo, Bunkyo-ku, Tokyo, 113-8656, Japan.
    {\texttt\small [kawaharazuka, miki, bando, t-suzuki, ribayashi, toshimitsu, nagamatsu, k-okada, inaba]@jsk.t.u-tokyo.ac.jp}
  }
}
\begin{document}

\maketitle
\thispagestyle{empty}
\pagestyle{empty}

\begin{abstract}
  \switchlanguage%
  {%
    Various musculoskeletal humanoids have been developed so far.
    While these humanoids have the advantage of their flexible and redundant bodies that mimic the human body, they are still far from being applied to real-world tasks.
    One of the reasons for this is the difficulty of bipedal walking in a flexible body.
    Thus, we developed a musculoskeletal wheeled robot, Musashi-W, by combining a wheeled base and musculoskeletal upper limbs for real-world applications.
    Also, we constructed its software system by combining static and dynamic body schema learning, reflex control, and visual recognition.
    We show that the hardware and software of Musashi-W can make the most of the advantages of the musculoskeletal upper limbs, through several tasks of cleaning by human teaching, carrying a heavy object considering muscle addition, and setting a table through dynamic cloth manipulation with variable stiffness.
  }%
  {%
    これまで様々な筋骨格ヒューマノイドが開発されてきた.
    これらは人体模倣型の柔軟かつ冗長な身体の利点を持つ一方で, 実世界へのアプリケーションへはまだまだ程遠い.
    この理由の一つに柔軟身体における二足歩行の困難性が挙げられ, 本研究ではメカナム台車と筋骨格上肢を合体させた筋骨格台車Musashi-Wを開発する.
    そして, 静的, または動的な身体図式を学習により獲得し, 反射制御, 視覚認識と合わせた全体ソフトウェアを構成する.
    これにより, 筋骨格台車Musashi-Wにおける教示による掃除動作, 筋追加を考慮した重量物体運搬実験, 柔軟布操作を含むテーブルセッティング実験に成功した.
  }%
\end{abstract}

\section{INTRODUCTION}\label{sec:introduction}
\begin{figure}[t]
  \centering
  \includegraphics[width=0.98\columnwidth]{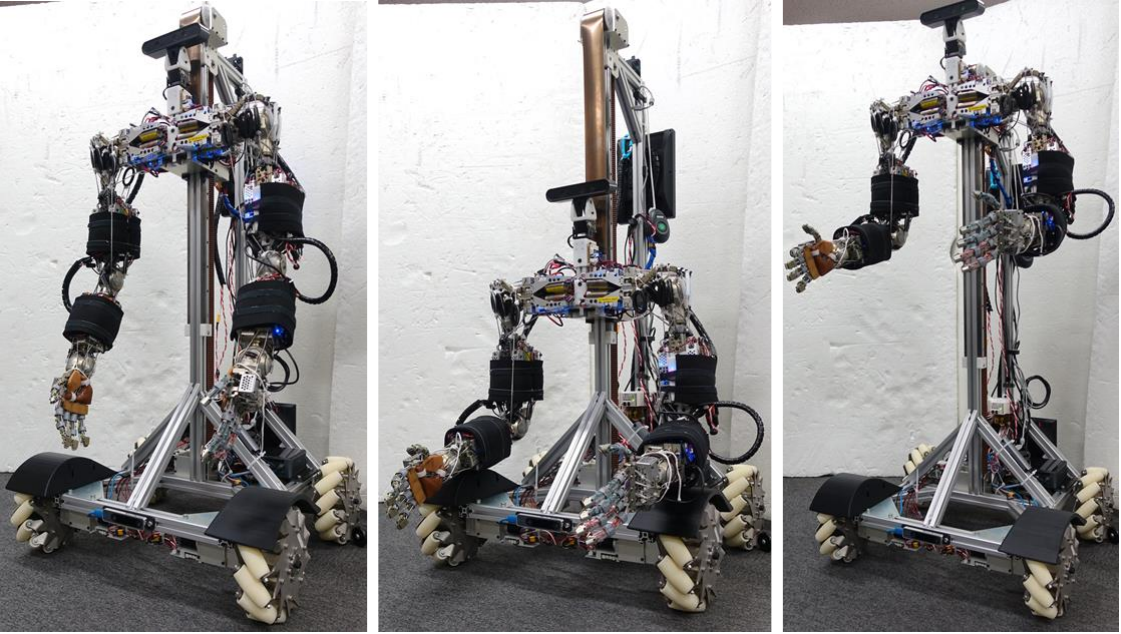}
  \caption{The newly developed musculoskeletal wheeled robot Musashi-W.}
  \label{figure:musashiw}
\end{figure}

\switchlanguage%
{%
  Various musculoskeletal humanoids have been developed so far \cite{gravato2010ecce1, jantsch2013anthrob, asano2016kengoro, kawaharazuka2019musashi}.
  These humanoids have structures that are characteristic of the human body, such as the spine structure \cite{osada2011planar}, the flexible five-fingered hand \cite{makino2018hand}, and the radioulnar structure of the forearm \cite{kawaharazuka2017forearm}.
  In addition, the redundancy of muscles and nonlinear elastic elements connected in series to the muscles allow for variable stiffness control \cite{kobayashi1998tendon}.
  The redundant muscles also allow for continuous motion in the case of muscle rupture \cite{kawaharazuka2022redundancy} and task-specific muscle addition \cite{kawaharazuka2022additional}.
  While these various studies have been conducted, these humanoids are far from being applied to real-world tasks.
  One of the reasons for this is the difficulty of bipedal walking in a flexible body.
  The flexible and complex bodies of musculoskeletal humanoids are difficult to control, and while various learning-based control methods have been proposed \cite{allessandro2013synergy, kawaharazuka2020autoencoder}, none have yet succeeded in walking control.
  Therefore, real-world applications that take advantage of the muscle redundancy, variable stiffness control, and various biomimetic features of the musculoskeletal structure have not been conducted.

  In this study, we develop a musculoskeletal wheeled robot, Musashi-W, which is a combination of musculoskeletal dual arms \cite{kawaharazuka2019musashi}, a mechanum-wheeled base, and a linear motion mechanism (\figref{figure:musashiw}) aiming at a more practical system.
  It is equipped with musculoskeletal upper limbs that include flexible five-fingered hands, redundant muscles, and nonlinear elastic elements, aimed to be applied to various real-world tasks.
  In addition, having a mechanum-wheeled base and a linear motion mechanism enables a wide range of stable movements to take advantage of the features of the musculoskeletal upper limbs.
  As for the software, in order to cope with the flexible and complicated hardware configurations, we constructed a learning system based on body schema learning \cite{kawaharazuka2020autoencoder, kawaharazuka2020dynamics}.
  Static body schema learning \cite{kawaharazuka2020autoencoder} is used for basic motion controls of the musculoskeletal body, and dynamic body schema learning \cite{kawaharazuka2020dynamics} is applied for more dynamic motions involving tools and target objects.
  In addition, we incorporate reflex controls to prevent the increase of internal force between antagonist muscles when performing a long series of motions \cite{kawaharazuka2020thermo, kawaharazuka2019relax}.
  We integrate them with visual recognition and classical controls to realize real-world applications that take advantage of the flexibility and redundancy of the musculoskeletal body.
}%
{%
  これまで様々な筋骨格ヒューマノイドが開発されてきた\cite{gravato2010ecce1, jantsch2013anthrob, asano2016kengoro, kawaharazuka2019musashi}.
  人間の身体に特徴的な背骨構造\cite{osada2011planar}や柔軟な五指ハンド\cite{makino2018hand}, 前腕の橈骨尺骨構造等を持つ\cite{kawaharazuka2017forearm}.
  また, 筋の冗長性と, 筋に直列に接続される非線形弾性要素により, 可変剛性制御が可能である\cite{kobayashi1998tendon}.
  冗長な筋肉により, 筋破断時の動作継続性\cite{kawaharazuka2022redundancy}や, タスクに応じた筋追加\cite{kawaharazuka2022additional}等も可能である.
  このように様々な研究が進んでいる一方で, 実世界へのアプリケーションには程遠い.
  この理由の一つに, 柔軟身体における二足歩行の困難性が挙げられる.
  筋骨格ヒューマノイドの柔軟で複雑な身体の制御は難しく, 様々な学習型制御が提案されてきている一方で\cite{allessandro2013synergy, kawaharazuka2020autoencoder}, 未だ歩行制御には成功した例がない.
  そのため, 柔軟で可変剛性制御可能かつ様々な生物規範型の利点を持つ身体の特徴を活かした実世界応用が行われてこなかった.

  そこで本研究では, より実用的なシステムを目指し, 筋骨格型の双腕\cite{kawaharazuka2019musashi}とメカナム4輪台車, 直動スライダーを合体させた, 筋骨格車輪Musashi-Wを開発する(\figref{figure:musashiw}).
  これは上半身に柔軟な五指ハンド, 冗長な筋肉, 非線形弾性要素を含むような筋骨格身体を備えることで, 様々な実世界タスクへと応用することを目指す.
  また, メカナム四輪台車と直動スライダーを有することで, 筋骨格上肢の利点を活かすだけの, 安定した幅広い動きが可能となる.
  一方で, ソフトウェアに関しては, 柔軟で複雑なハードウェア構成に対応するため, これまで開発してきた身体図式学習\cite{kawaharazuka2020autoencoder, kawaharazuka2020dynamics}をベースとした学習型システムを構築する.
  筋骨格身体の基本的な動作制御には静的身体図式学習\cite{kawaharazuka2020autoencoder}を用い, よりダイナミックな動きや物体・道具等を含む動作には動的身体図式学習\cite{kawaharazuka2020dynamics}を応用する.
  また, 長い一連の動作を行うにあたり, 拮抗筋間の内力上昇を防ぐ反射制御を組み込む\cite{kawaharazuka2020thermo, kawaharazuka2019relax}.
  その他, 視覚認識と台車制御を統合したシステムにより, 筋骨格身体の柔軟性と冗長性の利点を活かした実世界応用を実現する.
}%

\begin{figure*}[t]
  \centering
  \includegraphics[width=1.95\columnwidth]{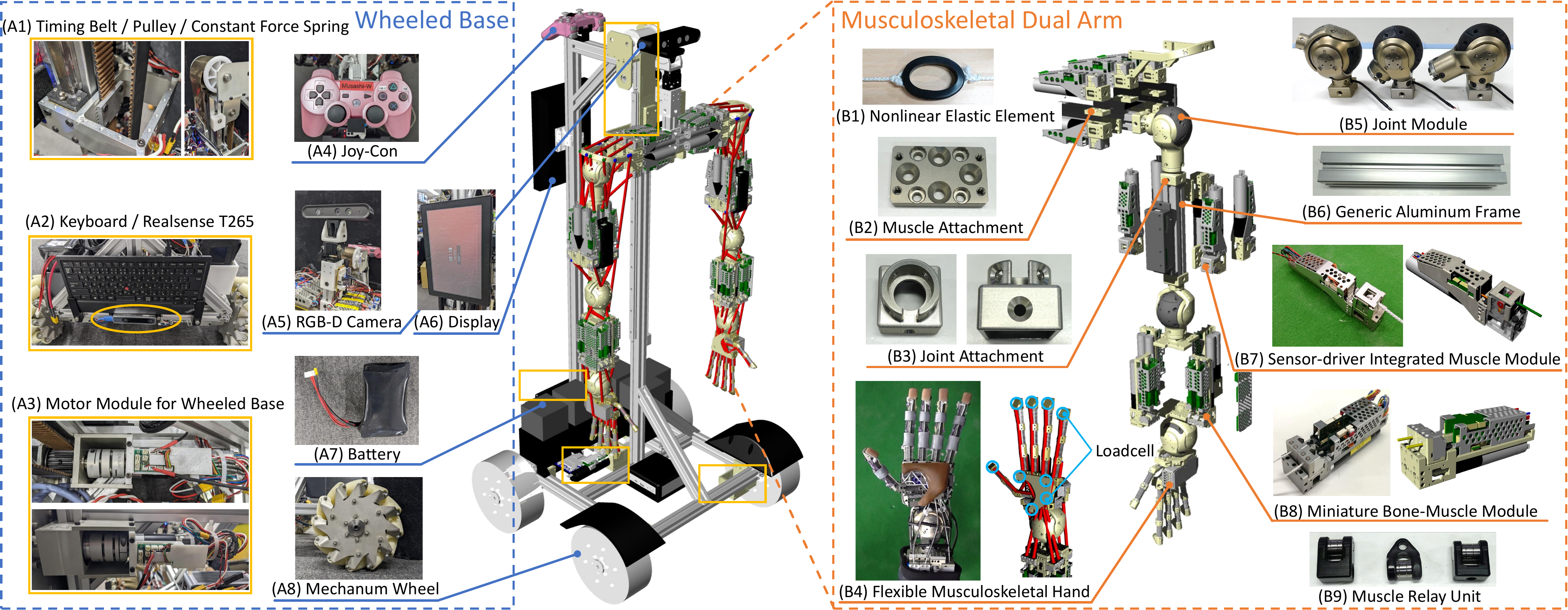}
  \caption{The hardware overview of Musashi-W.}
  \label{figure:hardware}
\end{figure*}

\switchlanguage%
{%
  The contributions of this study are as follows.
  \begin{itemize}
    \item Development of a musculoskeletal platform for real-world applications by integrating flexible and redundant musculoskeletal upper limbs with a wheeled base and a linear motion mechanism.
    \item Development of a learning-based software system that enables the realization of various movements by the developed complex body.
    \item Realization of real-world tasks using the developed hardware and software.
  \end{itemize}
  The structure of this study is as follows.
  In \secref{sec:hardware}, we describe the design of the musculoskeletal dual arms with various biomimetic features, the design of the wheeled base to enable a wide range of motion, and the overall design integrating these components.
  In \secref{sec:software}, we describe the overall software that integrates static and dynamic body schema learning, reflex control, visual recognition, etc. to operate the developed complex hardware.
  In \secref{sec:experiment}, we describe a cleaning experiment with human teaching, an object carrying experiment considering muscle addition, and a table setting experiment including dynamic cloth manipulation.
}%
{%
  本研究のコントリビューションは以下のようになっている.
  \begin{itemize}
    \item 柔軟かつ冗長な筋骨格上肢と車輪・直動機構を統合した実世界応用に向けた筋骨格プラットフォームの開発
    \item 開発した複雑な身体による動作実現を可能にする学習型ソフトウェアシステムの開発
    \item 開発したハードウェア・ソフトウェアによる実世界タスクの実現
  \end{itemize}
  本研究の構成は以下のようになっている.
  \secref{sec:hardware}では柔軟性と冗長性を持つ筋骨格双腕の設計, 幅広い動きを可能とする下肢設計, これらを統合した全体設計について述べる.
  \secref{sec:software}では開発した複雑なハードウェアを動作させる, 静的または動的な身体図式学習, 反射制御, 視覚制御等を統合した全体ソフトウェアについて述べる.
  \secref{sec:experiment}ではこれらハードウェアとソフトウェアを用いた人間の動作教示による掃除行動実験, 筋追加を考慮した重量物体運搬実験, 動的布操作を含むテーブルセッティング実験について述べる.
  最後に, \secref{sec:conclusion}において結論と今後の展望を述べる.
}%

\section{Hardware of Musculoskeletal Wheeled Robot Musashi-W} \label{sec:hardware}
\switchlanguage%
{%
  The hardware overview of the developed musculoskeletal wheeled robot Musashi-W is shown in \figref{figure:hardware}.
}%
{%
  開発した筋骨格台車Musashi-Wの全体像を\figref{figure:hardware}に示す.
}%
\subsection{Design of the Musculoskeletal Dual Arms} \label{subsec:musculoskeletal-arm}
\switchlanguage%
{%
  The detailed design of the musculoskeletal dual arms of Musashi-W is shown in the right figure of \figref{figure:hardware}.
  The basic skeletal structure consists of joint modules (B5), generic aluminum frames (B6), and joint attachments (B3).
  The joint module combines two types of central parts and three types of axial parts to enable various joint structures \cite{kawaharazuka2019musashi}.
  It contains joint angle sensors and IMU inside, which can be used as redundant sensors.
  The skeleton is formed by connecting the joint modules to generic aluminum frames using joint attachments.
  As muscle actuators, sensor-driver integrated muscle modules (B7) \cite{asano2015sensordriver} and miniature bone-muscle modules (B8) \cite{kawaharazuka2017forearm} are installed.
  The former is a muscle module that integrates a muscle motor, a motor driver, a muscle tension measurement unit, a pulley to wind the muscle, and a temperature sensor, to improve maintainability and reliability.
  The actuator is a Maxon BLDC motor 90W with 29:1 gear ratio, and 10 actuators are installed for each arm.
  In addition to the advantages of the former, the latter is a module that can be used as a skeleton by integrating two small motors and filling the space between them with metal.
  The actuators are Maxon BLDC motors 60W with 128:1 gear ratio, and eight actuators are installed for each forearm (four modules for each forearm).
  Both actuators can freely change the direction of the muscles depending on the placement of the muscle tension measurement units, and thus, together with muscle relay units (B9), various muscle arrangements can be realized.
  In addition, these muscle modules can be attached to generic aluminum frames or can be connected to each other by using muscle attachments (B2), making it easy to change the muscle arrangement or add new muscles.
  Since each muscle contains a nonlinear elastic element (B1), the body structure is flexible and its flexibility can be freely changed \cite{kawaharazuka2019longtime}.
  The fingers of the musculoskeletal hand (B4) are composed of machined springs, which provide a highly shock-resistant structure, and nine loadcells are placed on the fingertips and palm to detect contact \cite{makino2018hand}.
  The sensors mainly measure joint angle $\bm{\theta}$ from the joint module, muscle tension $\bm{f}$ from the muscle tension measurement unit, muscle length $\bm{l}$ from the encoder attached to the motor, motor temperature $\bm{c}$ from the temperature sensor, and contact force $\bm{F}$ from the loadcells in the hand.
}%
{%
  Musashi-Wの腕部分にあたる筋骨格双腕について述べる(\figref{figure:hardware}の右図).
  基本的な骨格構造は(B5)関節モジュールと(B6)汎用アルミフレーム, (B3)関節アタッチメントによって構成されている.
  関節モジュールは2種類の中心パーツと3種類の軸パーツを組み合わせることで様々な関節構造を可能にしている\cite{kawaharazuka2019musashi}.
  内部に関節角度センサとIMUを含み, 冗長なセンサとして利用できる.
  関節アタッチメントにより, 関節モジュールと汎用アルミフレームを接続することで骨格を成す.
  筋アクチュエータとして, (B7) sensor-driver integrated muscle module \cite{asano2015sensordriver}と(B8) miniature bone-muscle module \cite{kawaharazuka2017forearm}を搭載している.
  前者は, 筋モータ, モータドライバ, 筋張力測定ユニット, 筋を巻き取るプーリ, 温度センサ等を一体にし, メンテナンス性と信頼性を向上させた筋モジュールである.
  アクチュエータはMaxon BLDC motor 90W 29:1であり, 片腕に対して10本搭載されている.
  後者は, 前者の利点に加え, 小さな2つのモータを一体にし, その間を金属で満たすことで, 骨格としても利用可能なモジュールとなっている.
  アクチュエータはMaxon BLDC motor 60W 128:1であり, 片腕の前腕に対してを8本ずつ搭載されている(モジュールとしては4つである).
  どちらのアクチュエータも, 筋張力測定ユニットの配置次第で筋が出る方向を自由に変更することができるため, (B9)筋経由点ユニットと合わせて, 様々な筋配置を実現することができる.
  また, これらの筋モジュールは(B2)筋アタッチメントにより汎用アルミフレームに装着, または筋モジュール同士を接続させることができ, 筋配置変更や筋の新たな追加が容易な構成としている.
  それぞれの筋は(B1)非線形弾性要素を含むため, 柔軟な身体構造かつ, その柔軟性を自由に変化させることが可能である\cite{kawaharazuka2019longtime}.
  (B4)手の指は切削ばねによって構成されるため非常に衝撃に強い構造を持ち, 指先と手のひらに9つのロードセルを配置することで接触を検知することが可能である\cite{makino2018hand}.
  センサとしては主に, 関節モジュールから関節角度$\bm{\theta}$, 筋張力測定ユニットから筋張力$\bm{f}$, モータについたエンコーダから筋長$\bm{l}$, 温度センサからモータ温度$\bm{c}$, 手のロードセルから接触力$\bm{F}$を測定することができる.
}%

\subsection{Design of Wheeled Base} \label{subsec:wheeled-robot}
\switchlanguage%
{%
  Next, we describe the wheeled base and linear motion mechanism.
  The basic structure of the body is composed of the same generic aluminum frames as the musculoskeletal dual arms, and new components can be added and connected later.
  The wheels are four mechanum wheels 203 mm in diameter (A8), which allow the robot to move forward and backward, left and right, and rotate.
  A linear motion mechanism (A1) is constructed by two linear sliders, a timing belt, and a timing pulley.
  Musculoskeletal dual arms are connected to the linear motion mechanism, and a constant load spring with a load of 160 N is attached to reduce the load on the motor, so that little force is applied to the actuator when the robot is stationary.
  A motor (a Maxon BLDC motor 200W with 14:1 gear ratio), a motor driver, and a disc coupling attached to the end of the motor are combined into one motor module (A3), and the wheels and linear motion mechanism are driven by these five identical motor modules.
  The head is equipped with an RGB-D camera Astra S (A5, Orbbec 3D Technology International, Inc.).
  The neck is constructed by two servo motors, Dynamixel XM430-350 (ROBOTIS Co., Ltd.) in yaw and pitch axes.
}%
{%
  次に, 双腕部以外の台車と直動スライダー部について述べる.
  身体の基本構造は筋骨格双腕と同様の汎用アルミフレームにより構成されており, 新しいコンポーネントを後から好きに追加, 接続できるような構成としている.
  車輪については直径203 mmの(A8)メカナムホイールを4つ並べた形であり, 前後, 左右, 回転の動きを行うことができる.
  (A1)直動スライダー部については, リニアスライダー2つ, タイミングベルト, タイミングプーリを使い駆動している.
  直動部品に対して筋骨格双腕が接続するが, その重量を相殺してモータへの負荷を軽減するために, 荷重160 Nの定荷重バネを取り付けており, 静止時にはアクチュエータにはほとんど力がかからない.
  モータについては, Maxon BLDCモータ $\phi$30 200W 14:1を使用している.
  このモータとモータドライバ, モータの先についたカップリングを合わせて一つの(A3)モータモジュールとし, 全く同じモジュールを5つ使って車輪と直動が駆動されている.
  頭部にはRGBDカメラとして(A5) Atra S (Orbbec 3D Technology International, Inc.)が搭載されている.
  筋骨格双腕とカメラはyaw軸とpitch軸の2つのサーボモータDynamixel XM430-350 (ROBOTIS Co., Ltd.)によって接続されている.
}%

\begin{figure}[t]
  \centering
  \includegraphics[width=0.98\columnwidth]{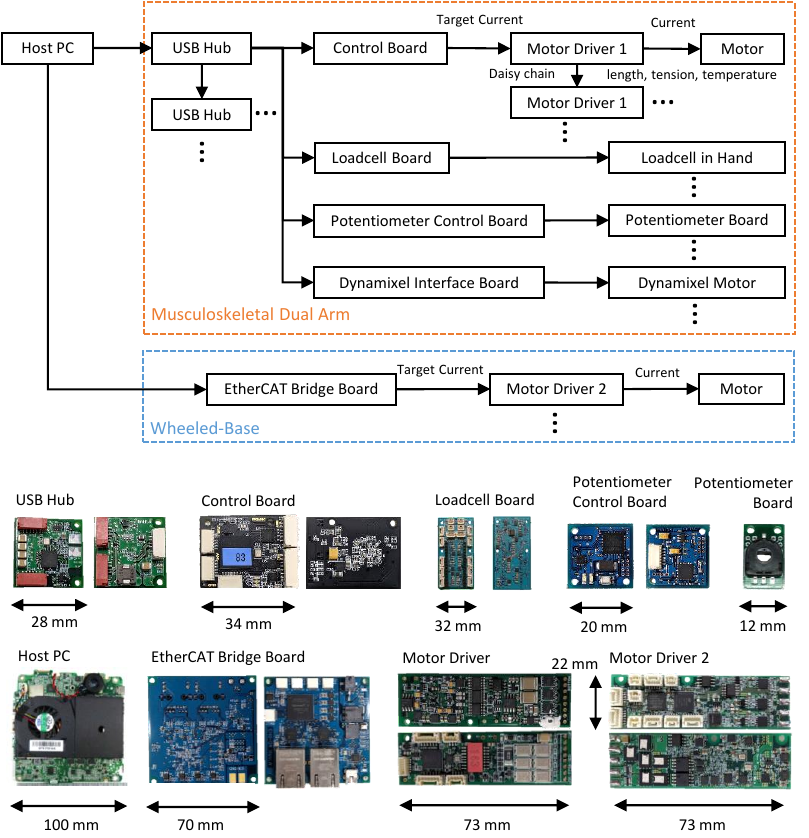}
  \caption{The circuit configuration of Musashi-W.}
  \label{figure:circuit}
\end{figure}

\begin{figure*}[t]
  \centering
  \includegraphics[width=1.7\columnwidth]{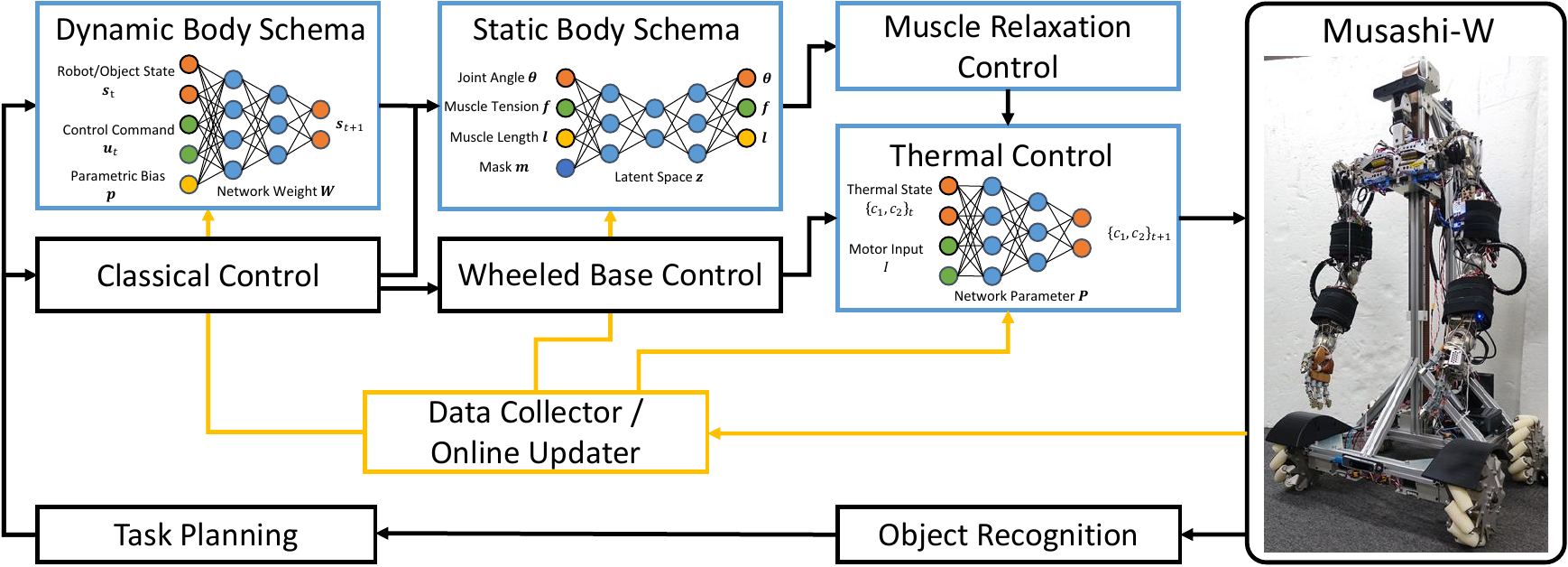}
  \caption{The software overview of Musashi-W.}
  \label{figure:software}
\end{figure*}

\subsection{Design of Musculoskeletal Wheeled Robot Musashi-W} \label{subsec:musashi-w}
\switchlanguage%
{%
  The musculoskeletal wheeled robot Musashi-W is a combination of the above musculoskeletal dual arms and the wheeled base.
  The height is 1.43 m at its maximum extension, and the weight is 55.2 kg including batteries.
  Musashi-W can carry up to five batteries with 13.2 V and 10000 mAh (A7), two of which can be connected in series to serve as logic power supplies, and three of which can be connected in series to serve as power supplies.
  PC is Intel NUC (Intel Corp.), and keyboard (A2) and touch-panel display (A6) are installed at the rear of the robot.
  In the lower part of the wheeled base, Realsense T265 (A2, Intel Corp.) is installed, which can perform visual SLAM.
  There is also a relay and a wireless emergency stop receiver in the middle, and a Joy-Con (A4) in the upper part.

  The circuit configuration is shown in \figref{figure:circuit}.
  The upper part of the body is based on USB communication, and Motor Driver 1 is connected from USB Hub through Control Board by a daisy chain.
  Serial communication is used from the Control Board onward.
  Potentiometer Control Board is placed in the joint module, and is connected to Potentiometer Board for each joint axis.
  As for the loadcells in the fingers and palm, Loadcell Board is placed on the back of the hand, and multiple loadcells are connected to it.
  Dynamixel motor on the neck is connected to Dynamixel Interface Board (DXHUB) by a daisy chain.
  Potentiometer Control Board, Loadcell Board, and Dynamixel Interface Board are connected to USB Hub for serial communication.
  On the other hand, the lower half of the body is based on Ethernet communication, and Motor Driver 2 is connected to EtherCAT Bridge Board by a daisy chain.
  Currently, the circuit configuration is different between the musculoskeletal dual arms and wheeled base, but since the size of the motor drivers (Motor Driver 1 and 2) has been unified, it is possible to adapt the entire configuration to the wheeled base side in the future.
}%
{%
  上記の筋骨格双腕と台車を合体させたのが, 筋骨格台車Musashi-Wである.
  身長は最大伸長時で143 cm, 体重はバッテリー含め55.2 kgである.
  (A7) バッテリーは13.2 V 10000 mAhのものを最大で5つ載せることができ, うち2つを直列に繋げてロジック用電源として, うち3つを直列に繋げてパワー用電源として使用することができる.
  PCはIntel NUC (Intel Corp.), ロボット後方には(A2)キーボードと(A6)タッチパネル式のディスプレイが搭載されている.
  台車下方には(A2) Realsense T265 (Intel, Corp.)が搭載され, Visual SLAMを行うことができる.
  また, 中部にはリレーと無線緊急停止のレシーバ, 上部には(A4) PS3コントローラが搭載されている.

  回路についてもその構成を\figref{figure:circuit}に示す.
  上半身は基本的にUSB通信を基本としており, USB HubからControl Boardを通して, その下にMotor Driver 1がdaisy chainで接続されている.
  Control Boardから先はシリアル通信が行われている.
  関節モジュール内にPotentiometer Control Boardが配置され, そこから関節軸ごとのPotentiometer Boardに接続している.
  指や手のひらのロードセルについては, 手の甲にLoadcell Boardが配置され, そこから複数のロードセルが接続されている.
  首のDynamixel Motorは, Dynamixel Interface Board (DXHUB)にdaisy chainで接続されている.
  このPotentiometer Control BoardやLoadcell Board, Dynamixel Interface BoardはUSB Hubと接続し, シリアル通信を行っている.
  一方で, 下半身はEthernet通信を基本としており, そこから直接モータドライバであるMotor Driver 2がdaisy chainで接続されている.
  現状筋骨格双腕と台車部で回路構成は異なるが, モータドライバのサイズを統一したため, 今後全て台車側の構成に合わせる予定である.
}%

\section{Software of Musculoskeletal Wheeled Robot Musashi-W} \label{sec:software}
\switchlanguage%
{%
  The entire system constructed in this study is shown in \figref{figure:software}.
  Basic motion control of the musculoskeletal body is performed by static body schema learning, and the dynamic body schema learning is responsible for dynamic motion control considering the information of tools and target objects.
  For the muscles in upper limbs, muscle relaxation control suppresses the increase in internal muscle force, and learning-based thermal control is performed for each motor.
  In addition, visual recognition, motion planning, inverse kinematics, and wheel control are integrated.
}%
{%
  本研究で構築した全体システムを\figref{figure:software}に示す.
  筋骨格身体の基本的な動作制御は静的身体図式学習により行い, その他の対象物体や環境等まで含めた動的な動作制御については動的身体図式が担う.
  上肢の筋については筋弛緩制御により内力上昇を抑え, それぞれのモータについて学習型の温度制御を行うという低レイヤの反射制御が働いている.
  その他, 認識や動作計画, 逆運動学や車輪制御については古典的な手法が働いている.
}%

\subsection{Static Body Schema Learning}
\switchlanguage%
{%
  Static body schema learning \cite{kawaharazuka2020autoencoder} is a learning mechanism for motion control of musculoskeletal bodies.
  By learning the relationship between joint angle $\bm{\theta}$, muscle tension $\bm{f}$, and muscle length $\bm{l}$, the robot is able to calculate the muscle length to achieve the desired joint angle and muscle tension, and to estimate unobservable joint angles.
  The static body schema can be expressed by the following equation,
  \begin{align}
    (\bm{\theta}, \bm{f}, \bm{l}) = \bm{h}_{static}(\bm{\theta}, \bm{f}, \bm{l}, \bm{m}) \label{eq:static}
  \end{align}
  where $\bm{h}_{static}$ is the static body schema represented by a neural network, and $\bm{m}$ represents a mask variable ($\in \{0, 1\}^3$).
  The network input is aggregated into a latent variable $\bm{z}$, and an output reproduces the input as in AutoEncoder \cite{hinton2006reducing}.
  For $\{\bm{\theta}, \bm{f}, \bm{l}\}$, each value can be inferred from the other two values, i.e., $(\bm{\theta}, \bm{f})\rightarrow\bm{l}$, $(\bm{f}, \bm{l})\rightarrow\bm{\theta}$, $(\bm{\theta}, \bm{l})\rightarrow\bm{f}$.
  In order to represent the mutual relationships in a single network of static body schema, we train it by changing $\bm{m}$ to three types: $\{\begin{pmatrix}1&1&0\end{pmatrix}^T, \begin{pmatrix}0&1&1\end{pmatrix}^T, \begin{pmatrix}1&0&1\end{pmatrix}^T\}$.
  The value of $\{0, 1\}$ represents the presence or absence of a mask for each value of $\{\bm{\theta}, \bm{f}, \bm{l}\}$.
  For example, if $\bm{m}=\begin{pmatrix}1&1&0\end{pmatrix}^T$, then the network input is $(\bm{\theta}, \bm{f}, \bm{0}, \begin{pmatrix}1&1&0\end{pmatrix}^T)$, where the corresponding value is masked while training.
  This static body schema is initially trained from the geometric model, and then it is learned online using the actual robot sensor information.
  For motion control, a loss function is defined that makes $\bm{\theta}^{pred}$ predicted from the static body schema close to the target value $\bm{\theta}^{ref}$ and minimizes the muscle tension $\bm{f}$.
  The target muscle length $\bm{l}^{ref}$ minimizing the loss is calculated by iteratively updating $\bm{z}$ through a backpropagation and gradient descent method.
  Also, variable stiffness control can be realized by adding a constraint to make the body stiffness close to the target value $\bm{k}^{ref}$.
  In addition, though Musashi-W has joint angle sensors, ordinary musculoskeletal humanoids do not have joint angle sensors \cite{asano2016kengoro}.
  In this case, data of $\bm{\theta}$ for learning can be calculated based on changes in muscle length and visual hand recognition \cite{kawaharazuka2018online}.
  By learning the static body schema, the robot can estimate $\bm{\theta}$ from the current $\bm{f}$ and $\bm{l}$ without continuously looking at the hand.
}%
{%
  静的身体図式学習\cite{kawaharazuka2020autoencoder}は筋骨格身体の動作制御を行うための学習機構である.
  関節角度$\bm{\theta}$, 筋張力$\bm{f}$, 筋長$\bm{l}$の間の相互関係を学習することで, 所望の関節角度や筋張力を実現するような筋長を計算したり, 観測できない関節角度を推定したりすることができるようになる.
  静的身体図式は以下のように式で表現できる.
  \begin{align}
    (\bm{\theta}, \bm{f}, \bm{l}) = \bm{h}_{static}(\bm{\theta}, \bm{f}, \bm{l}, \bm{m}) \label{eq:static}
  \end{align}
  ここで, $\bm{h}_{static}$はニューラルネットワークで表現された静的身体図式, $\bm{m}$はマスク変数($\in \{0, 1\}^3$)を表す.
  ネットワークの入力情報は潜在変数$\bm{z}$に集約され, その入力を再現するような出力を予測するAutoEncoder型のネットワークである.
  この$\{\bm{\theta}, \bm{f}, \bm{l}\}$の間には, それぞれの値が他2つの値から推論できる, つまり$(\bm{\theta}, \bm{f})\rightarrow\bm{l}$, $(\bm{f}, \bm{l})\rightarrow\bm{\theta}$, $(\bm{\theta}, \bm{l})\rightarrow\bm{f}$という三つ巴の関係が存在する.
  この相互関係の情報を静的身体図式の一つのネットワーク内で表現するため, $\bm{m}$を$\{(1 1 0)^T, (0 1 1)^T, (1 0 1)^T\}$の3種類に変化させながら学習を行う.
  $\{0, 1\}$の値は$\{\bm{\theta}, \bm{f}, \bm{l}\}$のそれぞれに関するマスクの有無を表している.
  例えば$\bm{m}=(1 1 0)^T$であれば, ネットワーク入力は$(\bm{\theta}, \bm{f}, \bm{0}, (1 1 0)^T)$のように, 対応する値をマスクしながら学習を行う.
  この静的身体図式を幾何モデルから初期学習し, その後実機データを使ってオンライン学習する.
  制御の際は, 静的身体図式から推論される$\bm{\theta}^{pred}$を指令値$\bm{\theta}^{ref}$に近づけ, 筋張力$\bm{f}$を最小化する損失関数を定義し, これをもとに$\bm{z}$を繰り返し更新することで, 指令を満たす指令筋長$\bm{l}^{ref}$が計算される.
  その他, 指令身体剛性$\bm{k}^{ref}$に近づけるような制約を加えることで, 可変剛性制御も実現することが可能である.
  また, Musashi-Wは関節角度センサを持つ一方で, 通常の筋骨格ヒューマノイドは関節角度センサを持たない\cite{asano2016kengoro}.
  この場合学習の際の$\bm{\theta}$のデータは筋長変化と視覚による手先認識をもとに行うことができる\cite{kawaharazuka2018online}.
  静的身体図式を学習することで, 常に手先を見なくとも, 現在の$\bm{f}$と$\bm{l}$から$\bm{\theta}$を推定することが可能である.
}%

\subsection{Dynamic Body Schema Learning}
\switchlanguage%
{%
  Dynamic body schema learning is a mechanism to learn a complex relationship among sensors and actuators related to the body, tools, and target objects, and to control them as intended.
  The dynamic body schema can be expressed by the following equation,
  \begin{align}
    \bm{s}_{t+1} = \bm{h}_{dynamic}(\bm{s}_{t}, \bm{u}_{t}, \bm{p}) \label{eq:dynamic}
  \end{align}
  where $\bm{s}$ represents the state of the robot and target object, $\bm{u}$ represents the control input, $\bm{h}_{dynamic}$ represents the dynamic body schema represented by a neural network, and $\bm{p}$ represents the parametric bias \cite{tani2002parametric}.
  $\bm{p}$ is a learnable input variable from which multiple attractor dynamics can be extracted; by collecting data while changing target objects and environments, information on the resulting changes in dynamics is embedded in $\bm{p}$.
  By setting a loss function that makes the prediction of the network close to the actual sensor information, and by updating only $\bm{p}$ while keeping the network weight $W$ fixed, the robot can recognize target objects and environments online and obtain dynamic body schema to match \cite{kawaharazuka2020dynamics}.
  For the motion control, the following calculation is repeated to obtain the optimized time-series control input $\bm{u}^{opt}_{seq}$,
  \begin{align}
    \bm{s}^{pred}_{seq} &= \bm{h}_{expand}(\bm{s}_{t}, \bm{u}^{opt}_{seq})\\
    L &= \bm{h}_{loss}(\bm{s}^{pred}_{seq}, \bm{u}^{opt}_{seq}) \label{eq:control-loss}\\
    \bm{u}^{opt}_{seq} &\gets \bm{u}^{opt}_{seq} - \gamma\partial{L}/\partial{\bm{u}^{opt}_{seq}} \label{eq:control-opt}
  \end{align}
  where $\bm{s}^{pred}_{seq}$ is the predicted time-series $\bm{s}$, $\bm{h}_{expand}$ is the time-series expansion of $\bm{h}_{dynamic}$, $\bm{h}_{loss}$ is the loss function, and $\gamma$ is the learning rate.
  In other words, the future $\bm{s}$ is predicted from $\bm{s}_{t}$ by $\bm{u}^{opt}_{seq}$, and $\bm{u}^{opt}_{seq}$ is optimized by the backpropagation and gradient descent methods to minimize the loss function set for the target task.
}%
{%
  動的身体図式学習は, 複雑で柔軟な身体内のセンサ・アクチュエータ, 道具, 対象物体等の間の関係性を学習することで, これらを意図したように制御する機構である.
  動的身体図式は以下のように式で表現できる.
  \begin{align}
    \bm{s}_{t+1} = \bm{h}_{dynamic}(\bm{s}_{t}, \bm{u}_{t}, \bm{p}) \label{eq:dynamic}
  \end{align}
  ここで, $\bm{s}$はロボットや対象物体等の状態, $\bm{u}$は制御入力, $\bm{h}_{dynamic}$はニューラルネットワークで表現された動的身体図式, $\bm{p}$はParametric Bias \cite{tani2002parametric}を表す.
  $\bm{p}$は複数のアトラクターダイナミクスを抽出できる学習可能な入力変数であり, 例えば様々な対象物体や環境でデータを取ることで, それらによるダイナミクス変化の情報が$\bm{p}$に埋め込まれる.
  ネットワークの予測と実測を合致させるような損失関数を設定し, ネットワーク重み$W$は固定した状態で$\bm{p}$のみ更新することで, オンラインで物体や環境の認識をし, それに合わせた動的身体図式を得ることができる\cite{kawaharazuka2020dynamics}.
  また, 制御の際には以下のような計算を繰り返すことで最適化された時系列の制御入力$\bm{u}^{opt}_{seq}$を得る.
  \begin{align}
    \bm{s}^{pred}_{seq} &= \bm{h}_{expand}(\bm{s}_{t}, \bm{u}^{opt}_{seq})\\
    L &= \bm{h}_{loss}(\bm{s}^{pred}_{seq}, \bm{u}^{opt}_{seq}) \label{eq:control-loss}\\
    \bm{u}^{opt}_{seq} &\gets \bm{u}^{opt}_{seq} - \gamma\partial{L}/\partial{\bm{u}^{opt}_{seq}} \label{eq:control-opt}
  \end{align}
  ここで, $\bm{s}^{pred}_{seq}$は予測された時系列の$\bm{s}$, $\bm{h}_{expand}$は$\bm{h}_{dynamic}$を時系列展開した関数, $\bm{h}_{loss}$は損失関数, $\gamma$は学習率を表す.
  つまり, $\bm{s}_{t}$から$\bm{u}^{opt}_{seq}$により将来の$\bm{s}$を予測し, これに対して設定した損失関数を最小化するように, $\bm{u}^{opt}_{seq}$を誤差逆伝播法と勾配降下法により最適化している.
}%

\subsection{Reflex Control}
\switchlanguage%
{%
  Muscle relaxation control and thermal control work as reflex controls.
  Muscle relaxation control \cite{kawaharazuka2019relax} is a reflex control that suppresses the increase in internal force due to modeling error, which is a problem due to the antagonistic relationship of the musculoskeletal structure.
  From the joint torque $\bm{\tau}^{nec}$ required to maintain the current posture, the necessary muscle tension $\bm{f}^{nec}$ is calculated by the quadratic programming method below,
  \begin{align}
    \textrm{minimize}&\;\;\;\bm{x}^{T}W_{1}\bm{x} + (G^{T}\bm{x}+\bm{\tau}^{nec})^{T}W_{2}(G^{T}\bm{x}+\bm{\tau}^{nec})\\
    \textrm{subject to}&\;\;\;\;\;\;\;\;\;\;\;\;\;\;\;\;\;\;\;\;\;\;\;\;\;\;\;\;\;\bm{x} \geq \bm{f}^{min}
  \end{align}
  where $G$ is the muscle Jacobian, $\bm{f}^{min}$ is the minimum muscle tension, and $\bm{f}^{nec}=\bm{x}$ for the calculated $\bm{x}$.
  Sorting $\bm{f}^{nec}$ in ascending order, the muscles are relaxed from the one with the lowest necessary muscle tension to an extent that does not affect the current posture.
  This method enables the robot to suppress the internal force in the antagonistic relationship and to perform continuous movements.
  Note that this reflex control cannot be used simultaneously with the variable stiffness control because it suppresses the internal force.

  Thermal control \cite{kawaharazuka2020thermo} is a reflex control that constantly learns the thermal model of the motor and simultaneously controls the temperature of the motor core to keep it within the rated value.
  We express the motor housing temperature as $c_{2}$, which can be measured, and the motor core temperature as $c_{1}$, which cannot be directly measured.
  A two-resistor thermal model of the motor is transformed as follows,
  \begin{align}
    \dot{c}_{1} = h_{1}(I, c_{1}, c_{2}) \label{eq:thermal-1}\\
    \dot{c}_{2} = h_{2}(c_{1}, c_{2}) \label{eq:thermal-2}
  \end{align}
  where $h_{1}$ and $h_{2}$ are functions represented by learnable parameters $\bm{P}$ set by humans, and $I$ represents the current to the motor.
  By setting $\bm{P}$ to a low dimension of 5 while referring to the thermal model of \equref{eq:thermal-1} and \equref{eq:thermal-2}, overfitting does not occur compared to a neural network, making it suitable for online learning.
  The motor core temperature $c_{1}$ can be estimated using $h_{1}$.
  $\bm{P}$ is updated online from the loss function that matches the prediction and actual measurement for $c_{2}$.
  Also, in the same way as in the dynamic body schema, the maximum current $I^{opt}$ can be calculated from the loss function that makes $c_{1}$ match the rated value.
  By limiting the current at this value, the motor core temperature can always be kept within the rated value.
  These two types of reflex controls enable long-term operation of the robot.
}%
{%
  反射制御として, 筋弛緩制御と温度制御が働いている.
  筋弛緩制御\cite{kawaharazuka2019relax}は, 筋骨格構造に特徴的な, 拮抗関係の問題点であるモデル化誤差による内力上昇を抑える反射制御である.
  現在の姿勢を保つのに必要な関節トルク$\bm{\tau}^{nec}$から, 現在必要と考えられる筋張力$\bm{f}^{nec}$を, 以下のような二次計画法を解くことで計算する.
  \begin{align}
    \textrm{minimize}&\;\;\;\bm{x}^{T}W_{1}\bm{x} + (G^{T}\bm{x}+\bm{\tau}^{nec})^{T}W_{2}(G^{T}\bm{x}+\bm{\tau}^{nec})\\
    \textrm{subject to}&\;\;\;\;\;\;\;\;\;\;\;\;\;\;\;\;\;\;\;\;\;\;\;\;\;\;\;\;\;\bm{x} \geq \bm{f}^{min}
  \end{align}
  ここで, $G$は筋長ヤコビアン, $\bm{f}^{min}$は最小筋張力を表し, 計算された$\bm{x}$について$\bm{f}^{nec}=\bm{x}$とする.
  $\bm{f}^{nec}$を昇順にソートし, 必要筋張力の低い筋から順に, 現在姿勢に影響を及ぼさない範囲で弛緩させていく.
  本手法により, 拮抗関係における筋内力を抑え継続的な動作が可能となる.
  一方で, 本反射は内力を抑制してしまうため, 可変剛性制御と同時に利用することはできない.

  温度制御\cite{kawaharazuka2020thermo}は, モータの温度モデルを常に学習すると同時に, モータコアの温度を定格内に保つように制御する反射制御である.
  測定可能なモータハウジング温度を$c_{2}$, 直接測定できないモータコア温度を$c_{1}$とし, モータの二抵抗温度モデルから以下のように式変形を施す.
  \begin{align}
    \dot{c}_{1} = h_{1}(I, c_{1}, c_{2}) \label{eq:5-thermal-proposed-3}\\
    \dot{c}_{2} = h_{2}(c_{1}, c_{2}) \label{eq:5-thermal-proposed-4}
  \end{align}
  なお, $h_{1}$と$h_{2}$は人間が設定した学習可能なパラメータ$\bm{P}$により表現された関数, $I$はモータへの電流を表す.
  モータの基本的な温度モデルを参考に$\bm{P}$を5次元の低次元で設定することで, ニューラルネットワークに比べ過学習しづらく, オンライン学習に適する.
  $c_{2}$に関する予測と実測を合致させる損失関数から$\bm{P}$をオンライン学習すると同時に, $h_{1}$を使いモータコア温度$c_{1}$を推定することができる.
  また, 動的身体図式の際と同様の形で, $c_{1}$を定格温度に合致させる損失関数から最大電流$I^{opt}$を最適化計算し, この値でモータに流す電流を制限することで, モータコア温度を定格内に常におさめることが可能になる.
  これら2つの反射制御により, 長期的な動作が可能となる.
}%

\begin{figure}[t]
  \centering
  \includegraphics[width=0.5\columnwidth]{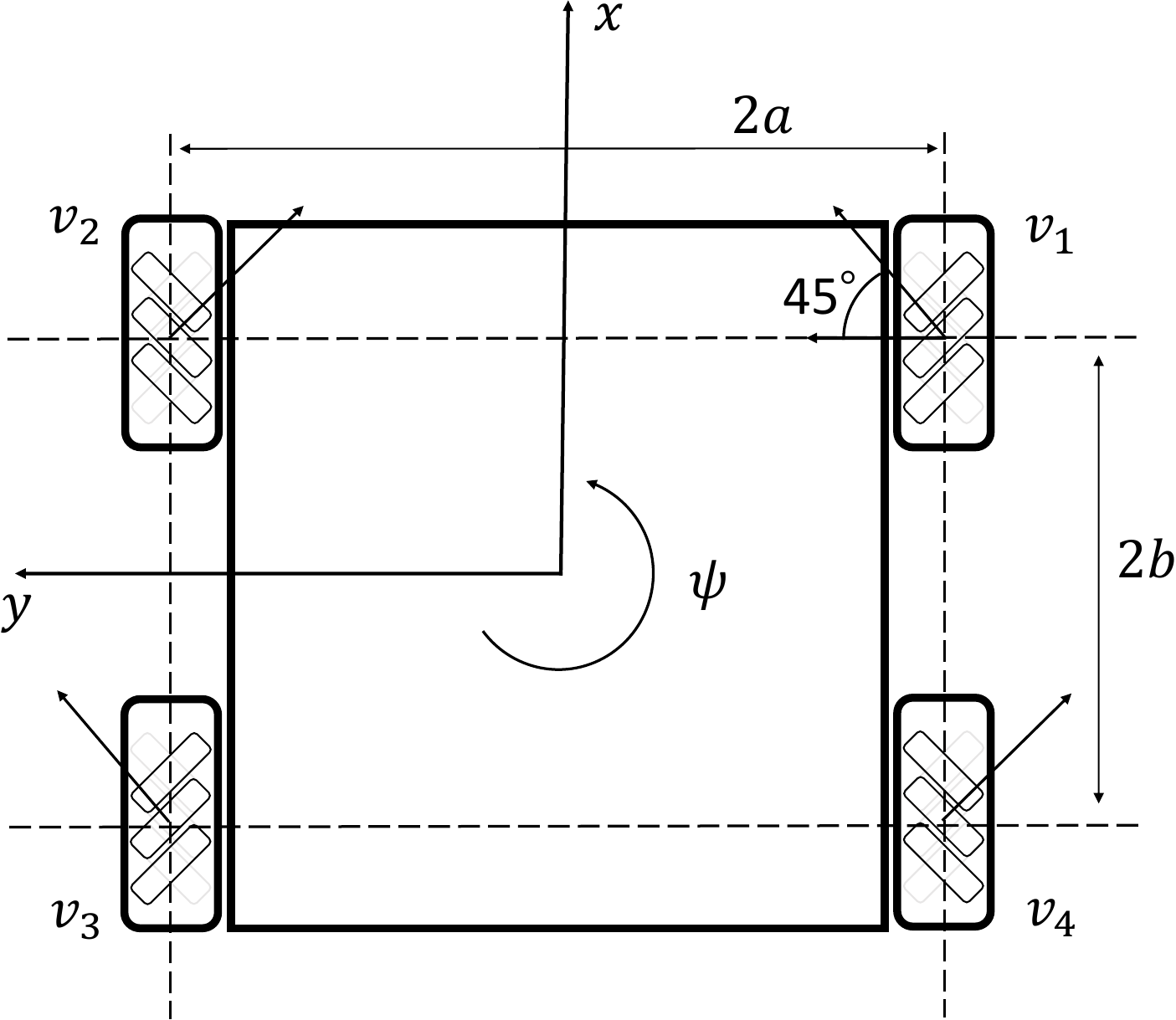}
  \caption{Parameter configuration of the wheeled base with four mechanum wheels.}
  \label{figure:wheel-control}
\end{figure}

\subsection{Visual Recognition and Classical Control}
\switchlanguage%
{%

  Other than the static body schema, dynamic body schema, and reflex controls described so far, we use general visual recognition, classical control, and task planning.
  Visual recognition is mainly based on point cloud recognition by color extraction and plane detection, and object recognition by color extraction and template matching.
  Due to its body flexibility, the musculoskeletal upper limbs do not require impedance control, but can grasp target objects by solving the inverse kinematics for the point cloud of the object, aligning the hand using the static body schema and visual feedback, and then sending a target value to press the hand against the object.
  Regarding the wheeled base, the velocity of each mechanum wheel is controlled.
  As shown in \figref{figure:wheel-control}, the distance between the wheels in the $y$-direction is $2a$, the distance between the wheels in the $x$-direction is $2b$, the speed of each wheel is $v_{\{1, 2, 3, 4\}}$, and the angle in the direction of rotation is $\psi$.
  The following equation holds for the wheel velocity and the translational and rotational velocities of the wheeled base:
  \begin{align}
    &\bm{v}_{wheel} = R\dot{\bm{x}}\\
    &\bm{v}_{wheel} = \begin{pmatrix}v_1\\v_2\\v_3\\v_4\end{pmatrix}, R = \begin{pmatrix}1&1&a+b\\1&-1&-(a+b)\\1&1&-(a+b)\\1&-1&a+b\end{pmatrix}, \dot{\bm{x}} = \begin{pmatrix}\dot{x}\\\dot{y}\\\dot{\psi}\end{pmatrix}
  \end{align}
  Therefore, for the target velocity $\dot{\bm{x}}$, by sending the velocity $R\dot{\bm{x}}$ to each wheel, it is possible to operate the robot in the desired direction.
  For odometry, it is possible to estimate the current $\bm{x}$ by repeating $\bm{x}\gets\bm{x} + R^{+}\bm{v}_{wheel}$.
}%
{%
  これまで説明した静的身体図式, 動的身体図式, 反射制御以外については, 一般的な視覚認識や古典的制御, タスク計画を用いている.
  視覚認識は主に色抽出や平面検出による点群の認識, 色抽出やテンプレートマッチングによる物体の認識を用いる.
  筋骨格上肢はその柔軟性ゆえにインピーダンス制御等を行う必要はなく, 物体の点群に対して逆運動学を解き, これに対して静的身体図式, またはこれを用いた視覚フィードバックを行って位置合わせをした後, 物体にめり込むように指令値を送ることで物体に馴染み把持することが可能である.
  台車制御はそれぞれのメカナム車輪の速度制御を行っている.
  \figref{figure:wheel-control}に示すように, $y$方向のタイヤ間距離を$2a$, $x$方向のタイヤ間距離を$2b$, それぞれのタイヤ速度を$v_{\{1, 2, 3, 4\}}$, 台車の回転方向の角度を$\psi$とする.
  車輪速度と台車の並進回転速度に以下の式が成り立つ.
  \begin{align}
    &\bm{v}_{wheel} = R\dot{\bm{x}}\\
    &\bm{v}_{wheel} = \begin{pmatrix}v_1\\v_2\\v_3\\v_4\end{pmatrix}, R = \begin{pmatrix}1&1&a+b\\1&-1&-(a+b)\\1&1&-(a+b)\\1&-1&a+b\end{pmatrix}, \dot{\bm{x}} = \begin{pmatrix}\dot{x}\\\dot{y}\\\dot{\psi}\end{pmatrix}
  \end{align}
  よって, 指令速度$\dot{\bm{x}}$に対して, $R\dot{\bm{x}}$を各車輪に速度指令することで, 所望の方向に動作することができる.
  また, オドメトリの際は$\bm{x} \gets \bm{x} + R^{+}\bm{v}_{wheel}$を繰り返すことで, 現在の$\bm{x}$を推定することが可能である.
}%

\section{Experiments} \label{sec:experiment}

\begin{figure}[t]
  \centering
  \includegraphics[width=0.98\columnwidth]{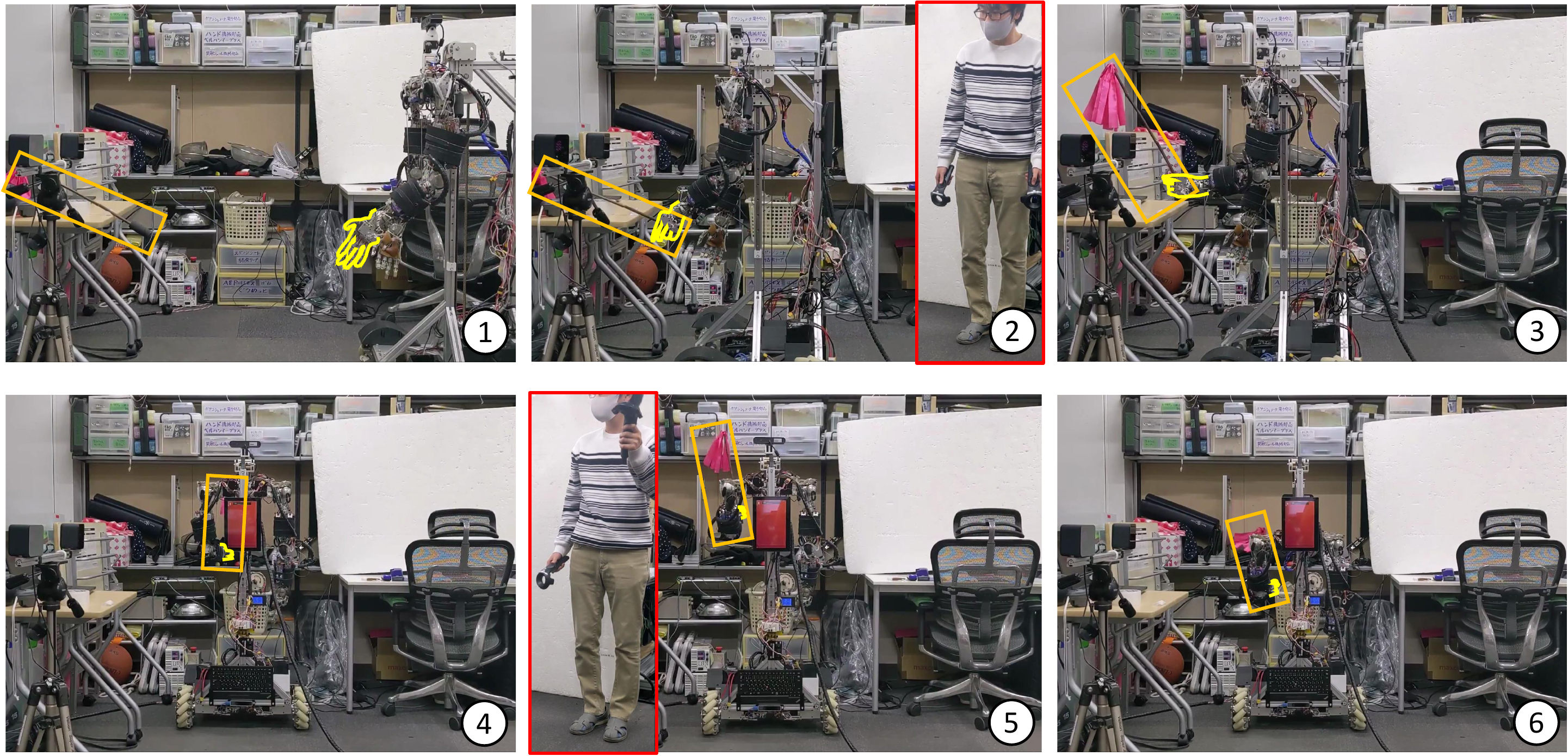}
  \caption{Duster experiment with human teaching.}
  \label{figure:human-demonstration}
\end{figure}

\subsection{Duster Experiment with Human Teaching}
\switchlanguage%
{%
  We describe a duster experiment operated by a VR controller which can move the wheeled base, linear motion mechanism, and musculoskeletal upper limbs.
  All software configurations except for the dynamic body schema are used here.
  We used HTC VIVE controller (HTC Corp.) and connected all actuators to the controller to construct a teaching system.
  While moving the robot hands from the position of the controller, the vertical movement of the linear joint and the translation and rotation of the wheeled base can be performed by operating the buttons on the controller.
  The experiment is shown in \figref{figure:human-demonstration}.
  During a period of about two minutes, the robot moves to the desk, picks up the duster, moves to the shelf, and successfully removes dust from a box on the shelf using the duster.
  We do not need to pay attention to environmental contact due to the flexibility of its body, and the robot can grasp various objects with its flexible five-fingered hand.
}%
{%
  台車や直動, 筋骨格上肢の全てをVRコントローラを使って動かし, はたき掃除を行う実験について述べる.
  ここでは動的身体図式以外の全てのソフトウェア構成が用いられている.
  HTC VIVEのコントローラを使い, 全てのアクチュエータをこれと接続して動作教示可能なシステムを構築した.
  コントローラの位置から手先を動かすと同時に, コントローラ上のボタン操作により直動関節の上下, 車輪の並進と回転が可能である.
  実際の実験の様子を\figref{figure:human-demonstration}に示す.
  約2分間の間に, 台車を進ませ, 机の上のはたきを取り, 棚まで移動して, 棚の上の箱のホコリを叩いて落とすことに成功している.
  身体の柔軟性から環境接触に対して気を配る必要がないと同時に, 柔軟な五指ハンドにより, 様々な物体を把持することができる.
}%

\begin{figure}[t]
  \centering
  \includegraphics[width=0.98\columnwidth]{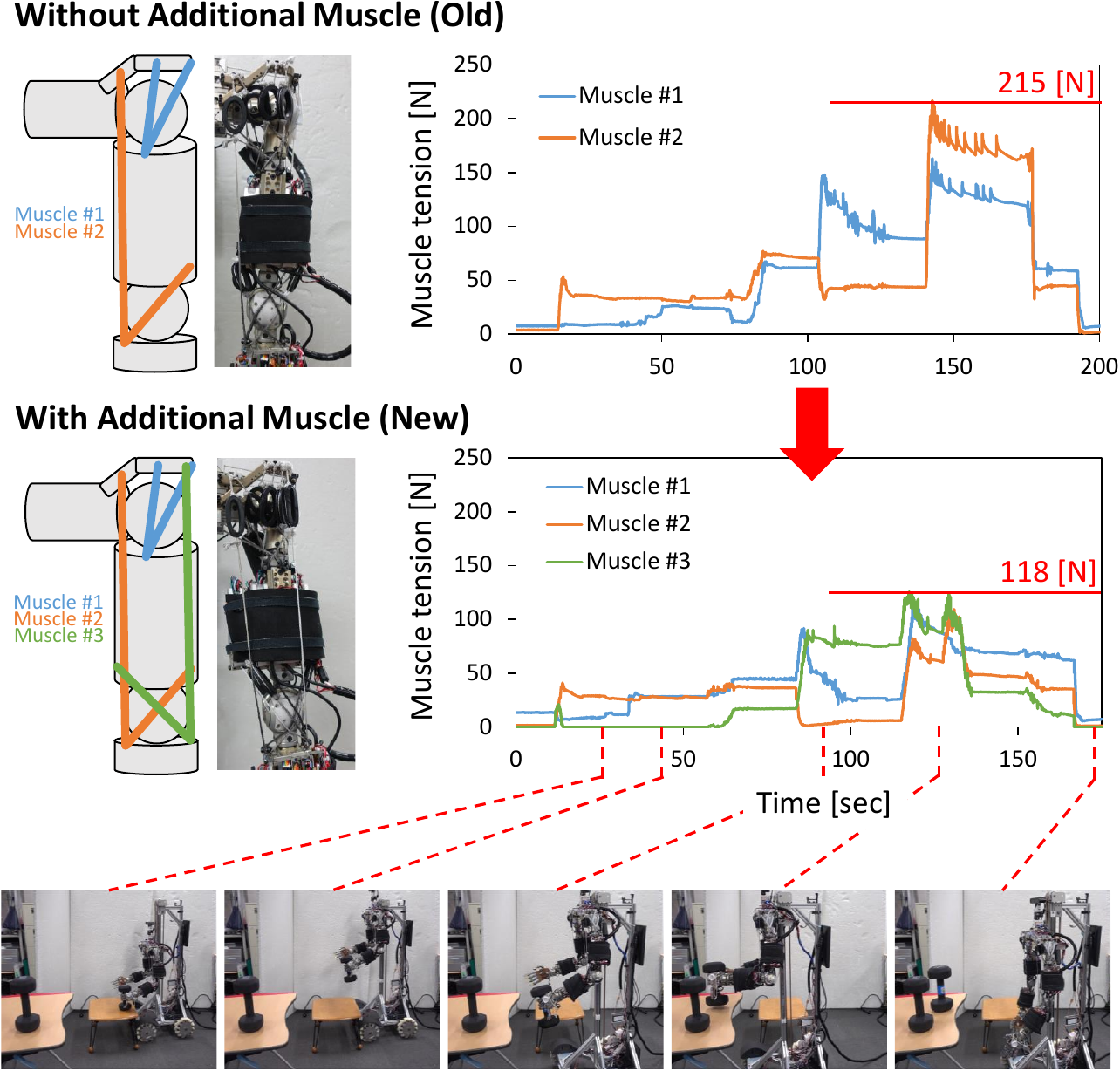}
  \caption{Object carrying experiment considering muscle addition.}
  \label{figure:heavy-object}
\end{figure}

\subsection{Object Carrying Experiment Considering Muscle Addition}
\switchlanguage%
{%
  We describe an experiment in which a task is performed by relearning a static body schema, using muscle redundancy, and adding muscles according to the task.
  All software configurations except the dynamic body schema are used here.
  As mentioned in \secref{subsec:musculoskeletal-arm}, muscle modules can be placed in various locations by means of muscle attachments, and we take full advantage of this.
  Note that a skilled researcher was able to complete the installation of new muscle attachments, muscle modules, wires, and cables in about three minutes.
  We conduct a task of carrying a heavy object about 6.8 kg in weight, as shown in the lower figures of \figref{figure:heavy-object}.
  In this case, a maximum force of 215N is applied during the operation with the usual muscle arrangement.
  On the other hand, by adding new muscles, increasing the input and output dimensions of the static body schema, and relearning it \cite{kawaharazuka2022additional}, the maximum muscle tension is reduced to 118N.
  In other words, by utilizing the advantages of the musculoskeletal body, the robot is able to perform real-world tasks while changing its body configuration according to the task.
}%
{%
  筋の冗長性を使い, タスクに応じて筋を追加して, 静的身体図式を再学習することでタスクを実行する実験について述べる.
  ここでは動的身体図式以外の全てのソフトウェア構成が用いられている.
  \secref{subsec:musculoskeletal-arm}でも述べたように, 筋骨格上肢は筋アタッチメントにより様々な場所に筋モジュールを配置できるため, この利点を最大限に活かす.
  なお, 筋アタッチメント, 筋モジュール, 筋ワイヤ, ケーブル類を新しく取り付けるのを, 熟練した研究者は約3分で終えることができた.
  \figref{figure:heavy-object}の下図のような, 約6.8kgの重量物体を運んで片付けるようなタスクを考える.
  このとき, 通常の筋配置であれば動作時に最大で215Nの力がかかってしまう.
  一方で, 新しく筋を追加し, 静的身体図式の入出力次元を増やし, これを再学習することで\cite{kawaharazuka2022additional}, 最大の筋張力が118Nまで減少する.
  つまり, 筋骨格上肢の利点を最大限活かすことで, タスクに応じて身体構成を変えながら実世界タスクをこなすことができるようになる.
}%

\begin{figure}[t]
  \centering
  \includegraphics[width=0.8\columnwidth]{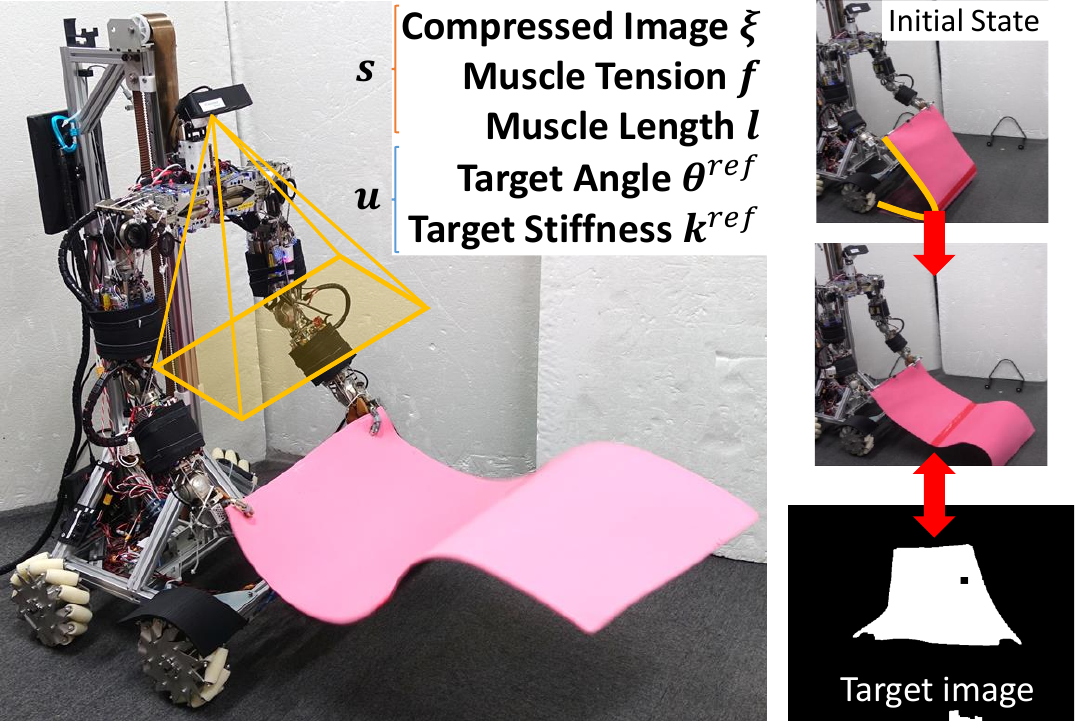}
  \caption{Dynamic cloth manipulation with variable stiffness.}
  \label{figure:cloth-manipulation}
\end{figure}

\begin{figure}[t]
  \centering
  \includegraphics[width=0.98\columnwidth]{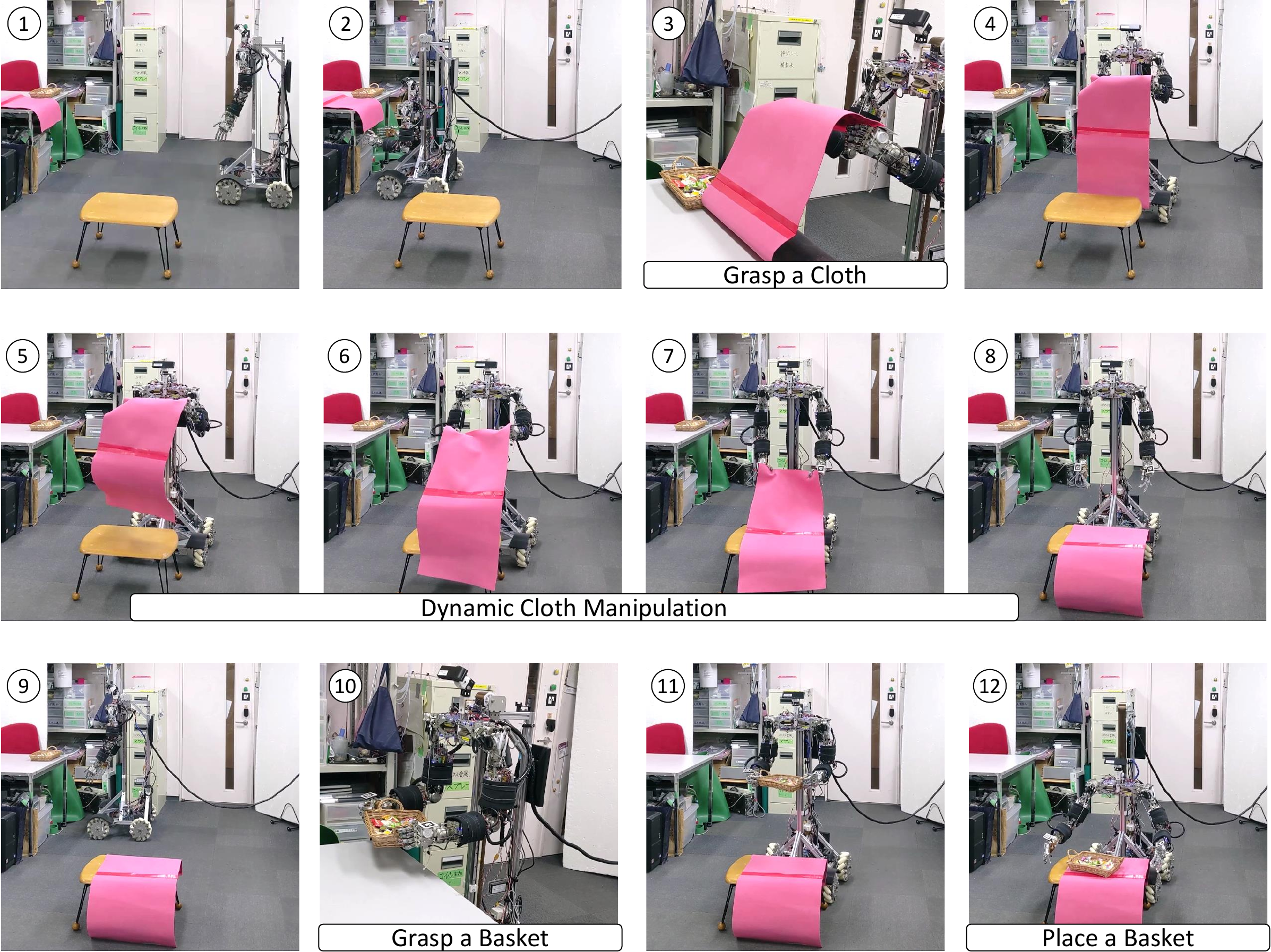}
  \caption{Table setting experiment with dynamic cloth manipulation.}
  \label{figure:table-setting}
\end{figure}

\subsection{Table Setting Experiment with Dynamic Cloth Manipulation}
\switchlanguage%
{%
  We describe a table setting experiment with dynamic cloth manipulation, which is enabled by learning dynamic body schema as well as utilizing the body's flexibility and variable stiffness.
  All software configurations including the dynamic body schema are used here.
  For the dynamic body schema, as shown in \figref{figure:cloth-manipulation}, the cloth image $\bm{\xi}$ compressed by AutoEncoder, muscle tension $\bm{f}$, and muscle length $\bm{l}$ are included in $\bm{s}$, and the target joint angle $\bm{\theta}$ and target body stiffness $\bm{k}^{ref}$ are included in $\bm{u}$.
  Using the image of an unfolded cloth as the target value, a control input $\bm{u}$ to achieve this image is obtained by optimization calculation \cite{kawaharazuka2022cloth}.
  A series of motions for table setting is shown in \figref{figure:table-setting}.
  The robot moves to the cloth on the desk and extracts the point cloud of the cloth using color extraction.
  The robot aligns its hand with the static body schema and visual feedback, grasps the cloth, and moves to the table.
  Using the dynamic body schema, the robot adjusts the transition of the target joint angle and body stiffness to unfold the cloth, and puts the cloth on the table.
  By manipulating the cloth while changing the body stiffness, the speed of the end effector is improved by about 12\% compared to without changing it.
  The robot moves to the basket of sweets on the desk again, and extracts the point cloud of the basket using plane recognition.
  The robot aligns its hand with the static body schema and visual feedback, grasps the basket, and moves to the table again.
  In this case, due to the flexibility of the arm, the robot is able to grasp the basket appropriately just by pressing its hand against it without control.
  Finally, the robot succeeds in placing the basket on the cloth covering the table.
  The entire task took about 7 minutes.
  By combining the musculoskeletal upper limbs with the wheeled base and linear motion mechanism, the robot has successfully performed a series of table setting operations by skillfully using the characteristics of the musculoskeletal structure, which have rarely been applied to real-world tasks.
}%
{%
  身体の柔軟性や可変剛性を利用すると同時に, 動的身体図式を学習することで動的布マニピュレーションを可能とし, 一連のテーブルセッティング動作を行う実験について述べる.
  ここでは動的身体図式を含む全てのソフトウェア構成が用いられている.
  動的身体図式については, \figref{figure:cloth-manipulation}に示すように, $\bm{s}$としてAutoEncoderにより圧縮された布の画像$\bm{\xi}$, 筋張力$\bm{f}$, 筋長$\bm{l}$を用い, $\bm{u}$として指令関節角度$\bm{\theta}^{ref}$と指令剛性$\bm{k}^{ref}$を用いる.
  布を広げたような画像を指令値として, これを実現するような$\bm{u}$を最適化計算により求める\cite{kawaharazuka2022cloth}.
  一連の動作実験の様子を\figref{figure:table-setting}に示す.
  机の上の布まで移動し, 色抽出から布の点群を取り出す.
  これに対して静的身体図式と視覚フィードバックにより手の位置を合わせこれを把持し, テーブルの上まで移動する.
  動的身体図式を使い, 身体の指令関節角度と指令剛性の遷移を調整することで, 布を広げて机の上にかぶせる.
  なお, 身体剛性を変化させながら操作することで, 手先速度は約12\%向上した.
  もう一度机の上にあるお菓子の入ったバスケットまで移動し, 平面認識から物体の点群を切り出す.
  これに対して静的身体図式と視覚フィードバックにより手の位置を合わせこれを把持し, またテーブルの位置まで移動する.
  この際, 腕の柔軟性から, 制御なしに適当に手を押し付けるだけでバスケットを適切に把持することができた.
  最後に, このバスケットをテーブルにかぶせた布の上に置くことに成功した.
  全体のタスク時間は約7分である.
  筋骨格上肢と台車・直動を合わせることで, これまで実世界タスクにほとんど応用されてこなかった筋骨格上肢を巧みに用いた一連動作に成功している.
}%

\section{CONCLUSION} \label{sec:conclusion}
\switchlanguage%
{%
  In this study, we developed a musculoskeletal wheeled robot, Musashi-W, which is a combination of musculoskeletal dual arms and a mechanum-wheeled base, and constructed a learning-based software system for real-world applications by utilizing the advantages of the musculoskeletal body.
  The musculoskeletal body has a mechanism that can adaptively change its flexibility by using nonlinear elastic elements, and it has the characteristics of redundant muscles and modularity for easy muscle addition.
  Because of the complexity of the body, it is necessary to learn a static relationship between its sensors and actuators and a dynamic relationship between the body, tools, and target objects.
  In addition, reflex controls to suppress the increase in internal force caused by redundant muscles, and visual feedback control based on static body schema are important.
  As a result of these efforts, the musculoskeletal wheeled robot Musashi-W succeeded in carrying heavy objects while considering muscle addition and performing a series of table-setting operations including dynamic cloth manipulation and object grasping.
  It is possible to solve real-world tasks by taking advantage of the muscle redundancy, body flexibility, and variable stiffness of the upper limbs, and by combining the wheeled base with linear motion mechanism.
  We believe that the hardware configuration with a flexible body and the online learning-based software configuration proposed in this study will be important elements in performing real-world tasks.
}%
{%
  本研究では, 筋骨格身体の利点を活かした実世界応用に向けて, 筋骨格型の双腕とメカナム型の台車を合体させた筋骨格台車Musashi-Wを開発し, その学習型全体システムを構成した.
  筋骨格型の身体は非線形弾性要素により柔軟性を適応的に変更可能な機構を持ち, 冗長な筋肉の特性とモジュラー性を活かした容易な筋追加機能を持つ.
  その身体は非常に複雑であるため静的な身体感覚間の関係や, 物体との間の動的な関係を学習する必要があった.
  また, 冗長な筋肉に起因する内力上昇を抑える反射制御や, 静的身体図式を応用した視覚フィードバックが重要である.
  これらにより, 筋骨格台車Musashi-Wが筋追加を考慮した重量物体運搬や, 動的布操作や物体把持を含む一連のテーブルセッティング動作等に成功した.
  上肢の筋冗長性や身体柔軟性, 可変剛性を活かしつつ, 台車と直動を合わせることで, 実世界タスクを新しい形で解くことができる.
  本研究で提案した柔軟身体を持つハードウェア構成とオンライン学習型のソフトウェア構成は今後実世界タスクを行う上で非常に重要な要素となると考える.
}%

{
  \bibliographystyle{IEEEtran}
  \bibliography{bib}
}

\end{document}